\documentclass[letterpaper, 10 pt, conference]{ieeeconf}
\usepackage{amsmath,amsfonts}
\usepackage{algorithmic}
\usepackage{algorithm}
\usepackage{array}
\usepackage[caption=false,font=normalsize,labelfont=sf,textfont=sf]{subfig}
\usepackage{textcomp}
\usepackage{url}
\usepackage{verbatim}
\usepackage{graphicx}
\usepackage{multirow}
\usepackage{cite}
\usepackage{booktabs}
\usepackage[hidelinks]{hyperref}

\usepackage[table]{xcolor}
\definecolor{lightgray}{gray}{0.9}
\newcommand{\highlight}[1]{\cellcolor{lightgray}#1}
\usepackage{placeins}
\newcolumntype{C}[1]{>{\centering\arraybackslash}m{#1}}
\usepackage[font=footnotesize,labelsep=colon]{caption}

\begin{document}

\title{\LARGE \bf DeeperRadar: End-to-End MIMO Radar Design and Multi-Modal Fusion for Autonomous Vehicle Perception}

\author{%
Eli Goldenshluger$^{1}$, Barak Pinkovich$^{1}$, and Chaim Baskin$^{2}$%
\\[0.25em]
{\footnotesize
$^{1}$Technion--Israel Institute of Technology
\qquad
$^{2}$Ben-Gurion University of the Negev}%
}


\maketitle
\thispagestyle{empty}
\pagestyle{empty}

\begin{abstract}
DeeperRadar is a radar-centric, sensor-stack-conditioned framework that co-designs radar sensing and multi-modal 3D detection for autonomous mobility by learning a sparse acquisition pattern end-to-end with the fusion model. As illustrated in Fig.~\ref{fig:intro_pipeline}, a learnable MIMO design module is trained end-to-end within a fusion network that operates directly on raw radar ADC data together with camera images and LiDAR point clouds. During training, the design module is supervised by the other sensors, enabling the system to learn both which receiver antennas to activate and the effective number of them. At deployment, the design module is removed and replaced by the learned sparse subsampling mask, leaving the downstream model architecture unchanged. Evaluated on the RADIal dataset, DeeperRadar discovers sparse, task-aware radar configurations that match or exceed full-array baselines while using fewer receivers, potentially reducing radar cost and integration complexity. These results show that learned optimal MIMO radar design depends on the fusion stack and the downstream perception task. Project page at: https://egoldensh.github.io/DeepeRadar.
\end{abstract}

\section{INTRODUCTION}
Mobile robots and autonomous vehicles rely on a rich perception stack that fuses information from multiple complementary sensors, such as cameras, LiDAR, and radar~\cite{drews2022deepfusion}. Cameras passively sense visible light and produce dense RGB images that are well suited for object recognition and scene understanding. LiDAR actively measures time-of-flight and returns 3D point clouds, enabling accurate geometric reasoning and the detection of small objects. However, both cameras and LiDAR can degrade under challenging conditions such as glare, low contrast, fog, dust, or heavy rain. Moreover, a camera cannot physically provide range measurements, and neither cameras nor conventional automotive LiDAR can provide direct velocity measurements. LiDAR-based velocity sensing remains largely experimental and is not yet widely deployed \cite{LidarVel2021}.
\par
In contrast, radar, a radio-frequency sensor operating at millimeter-wave frequencies, directly measures radial velocity via Doppler shifts and remains effective in many of these adverse conditions \cite{vargas2021radarweak}. Millimeter-wave multiple-input multiple-output (MIMO) radar is a well--developed sensing technology. It relies on an array of transmit (Tx) and receive (Rx) antennas and enables high-resolution perception \cite{fishler2004mimo}. In practice, modern automotive MIMO systems sample a virtual Tx–Rx array and, after standard range–angle–Doppler (RAD) processing and target extraction, produce a sparse 3D radar point cloud \cite{richards2014fundamentals}. 
\begin{figure}[!htbp]
    \centering
    \includegraphics[width=\linewidth]{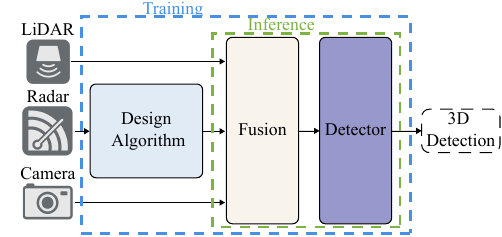}
    
    \caption{High-level overview of our radar-centric pipeline: during training, LiDAR, radar, and camera supervise a learnable MIMO design module that learns a subsampling mask on raw radar data; at deployment, the learned mask is applied to the radar, which is fused with LiDAR and camera features for 3D detection.}
    \label{fig:intro_pipeline}
\end{figure}
To achieve higher angular resolution, a large number of Rx channels is required (see Fig.~\ref{fig:intro_triptych}), which expands the virtual aperture but also increases cost and integration complexity. Compressed sensing (CS) methods \cite{CS2006} can mitigate these costs by enabling fewer samples and/or channels while preserving fidelity \cite{Yu_2010,Rossi_2014}. Complementary to these model-based approaches, recent work proposes some learnable MIMO design frameworks. Such systems, for a fixed Rx channel budget, jointly optimize in a self--supervised manner which subset of channels to retain and how to reconstruct the range–angle image. It is reported in \cite{weiss2021mimoOptim} that learnable MIMO designs achieve higher quality than conventional deterministic layouts or random configurations. However, by remaining radar-only and agnostic to the sensor stack, they have not been used in the context of multi–sensor fusion. This raises a key question: is MIMO-radar design determined solely by its hardware and the required end task, or also by the sensors that are fused with it?
\par
Multi-sensor fusion typically combines raw/low-level features, intermediate representations, or final detections. Classical systems emphasize late fusion at the detection level, whereas recent deep learning approaches operate on raw or low-level features to preserve geometry, phase, and Doppler cues\cite{fusionreview2025}. In particular, Bird’s-Eye-View (BEV) has become the main method that unifies multi-view camera features and, when available, additional modalities into a common 3D space that supports efficient detection, occupancy, and mapping \cite{liu2024bevfusion,ge2023metabev,Nabati_2021,yang2024ralibevradarlidarbev,bevguide2023}. Among these approaches, EchoFusion \cite{liu2023echoes} bypasses point-cloud sparsification and fuses raw radar with images in BEV, yielding substantial gains. This is achieved via a polar-aligned attention mechanism that associates image columns with radar range bins, effectively assigning radar the role of range sensing while leveraging the camera for angular and semantic cues.
\par
In this work, building on the above observations, we consider the impact of sub-Nyquist receiver sampling on angular resolution and ask the following question.
Is it possible to deliberately trade radar angular resolution for reduced MIMO complexity, recover the missing angular detail from the camera relying on radar primarily for range, and at the same time maintain or even improve 3D detection performance with
fewer Rx channels?
\par
We present a radar-centric, task-conditioned, end-to-end design-and-fusion framework that learns both the radar configuration and the fusion pipeline. Our approach jointly learns receiver selection, fusion, and detection in a single training loop, improving upon radar-only methods by learning not only which predefined number of discrete Rx antennas to activate but also the effective Rx budget via a fractional, differentiable Top-K selector (extending \cite{weiss2021mimoOptim}). Built atop a BEV-based polar aligned multi-modal architecture inspired by \cite{liu2023echoes}, the proposed radar design is co-optimized alongside image and LiDAR features and adapts to the chosen modality combination. Across extensive experiments on both simulation and real-world data, we show that the learned configurations achieve higher mean performance than fixed or random Rx layouts. In addition, the proposed design yields comparable or better overall 3D detection accuracy, supporting the hypothesis that optimal MIMO design depends on the fusion context. Using raw radar data jointly with image and LiDAR data from the RADIal dataset \cite{rebut2022}, we further demonstrate that our approach achieves competitive 3D detection accuracy relative to EchoFusion while substantially reducing MIMO radar complexity. To the best of our knowledge, RADIal is the only publicly available multi-modal dataset that provides synchronized raw ADC radar measurements from a MIMO radar, along with LiDAR and camera data, and the necessary radar steering matrix. Thus, this dataset is well-suited to evaluating our method.
\par
Our main contributions are summarized as follows:
\begin{itemize}
    \item \textbf{Budget-aware MIMO design.} We extend the MIMO design module to explicitly incorporate budget learning, enabling the network to jointly learn the receiver subset and budget.
    \item \textbf{End-to-end MIMO design and fusion.} We integrate radar design end-to-end with multi-modal fusion for radar–camera, radar–LiDAR, and radar-camera-LiDAR stacks, achieving comparable or higher observed 3D accuracy with reduced radar complexity relative to the baseline.
    \item \textbf{Stack--dependent radar analysis.} We provide quantitative and qualitative analyses showing that optimal radar layouts vary with the sensor suite and downstream task.
\end{itemize}

Empirically, we find that reducing MIMO radar complexity does not necessarily degrade downstream performance in radar–camera fusion and can sometimes improve it. Because smaller radar arrays yield lower angular resolution, the model relies on cameras for superior angular cues and uses radar for range. This is consistent with the column-wise image–row-wise radar alignment in our polar fusion method.

\par
\begin{figure}[t]
    \centering
    
    \includegraphics[width=0.32\linewidth,keepaspectratio]{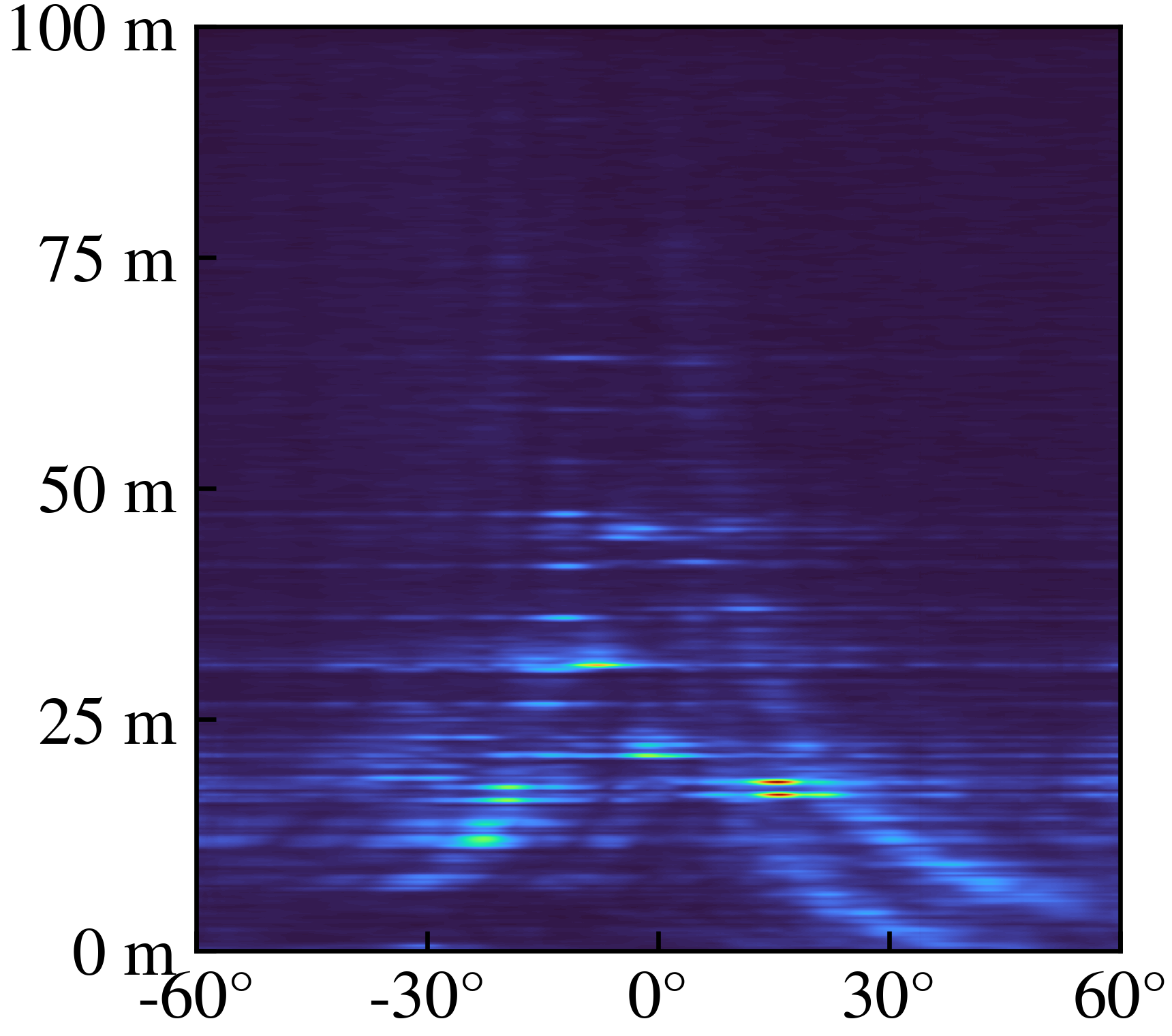}
    \includegraphics[width=0.32\linewidth,keepaspectratio]{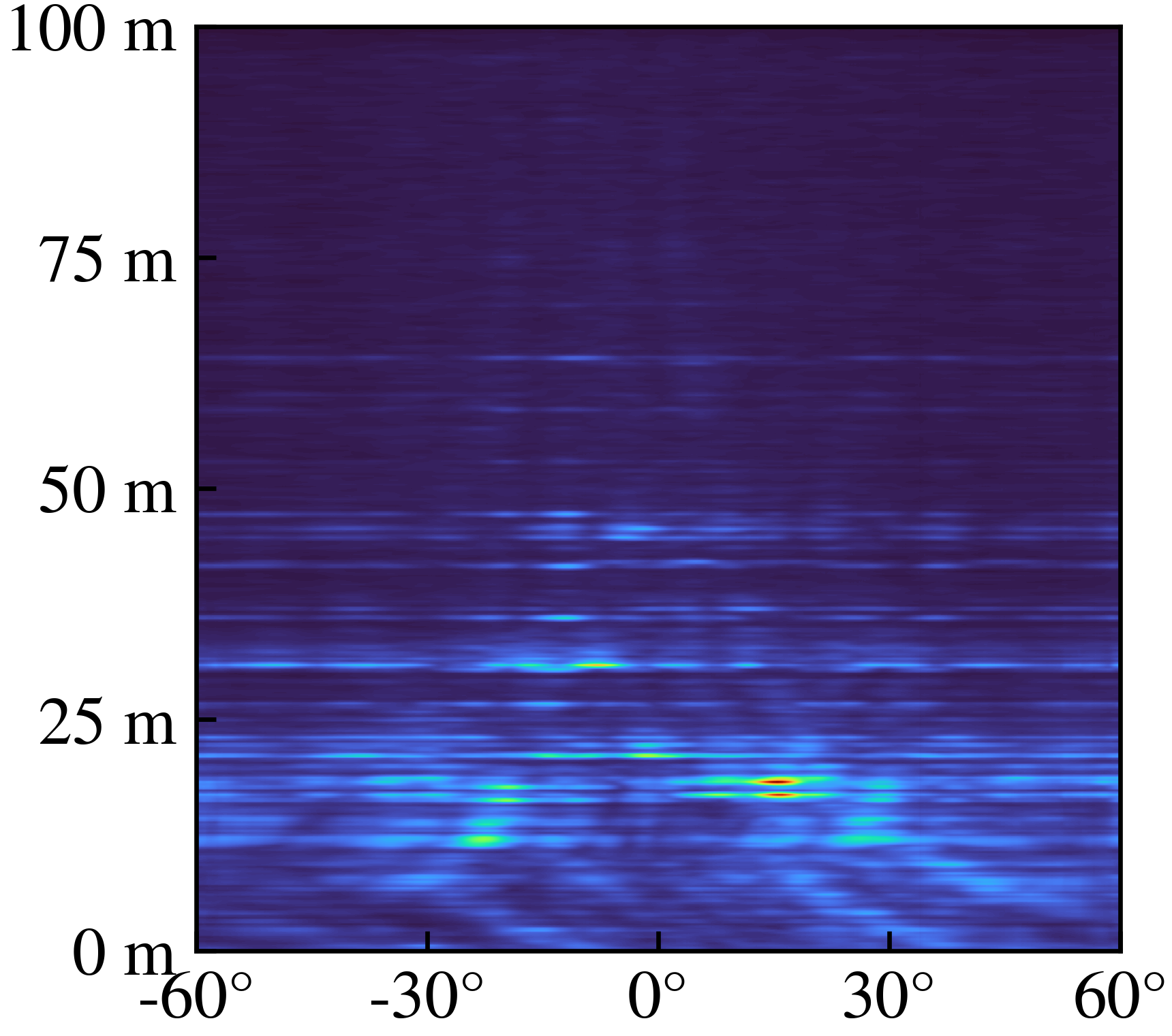}
    \includegraphics[width=0.32\linewidth,keepaspectratio]{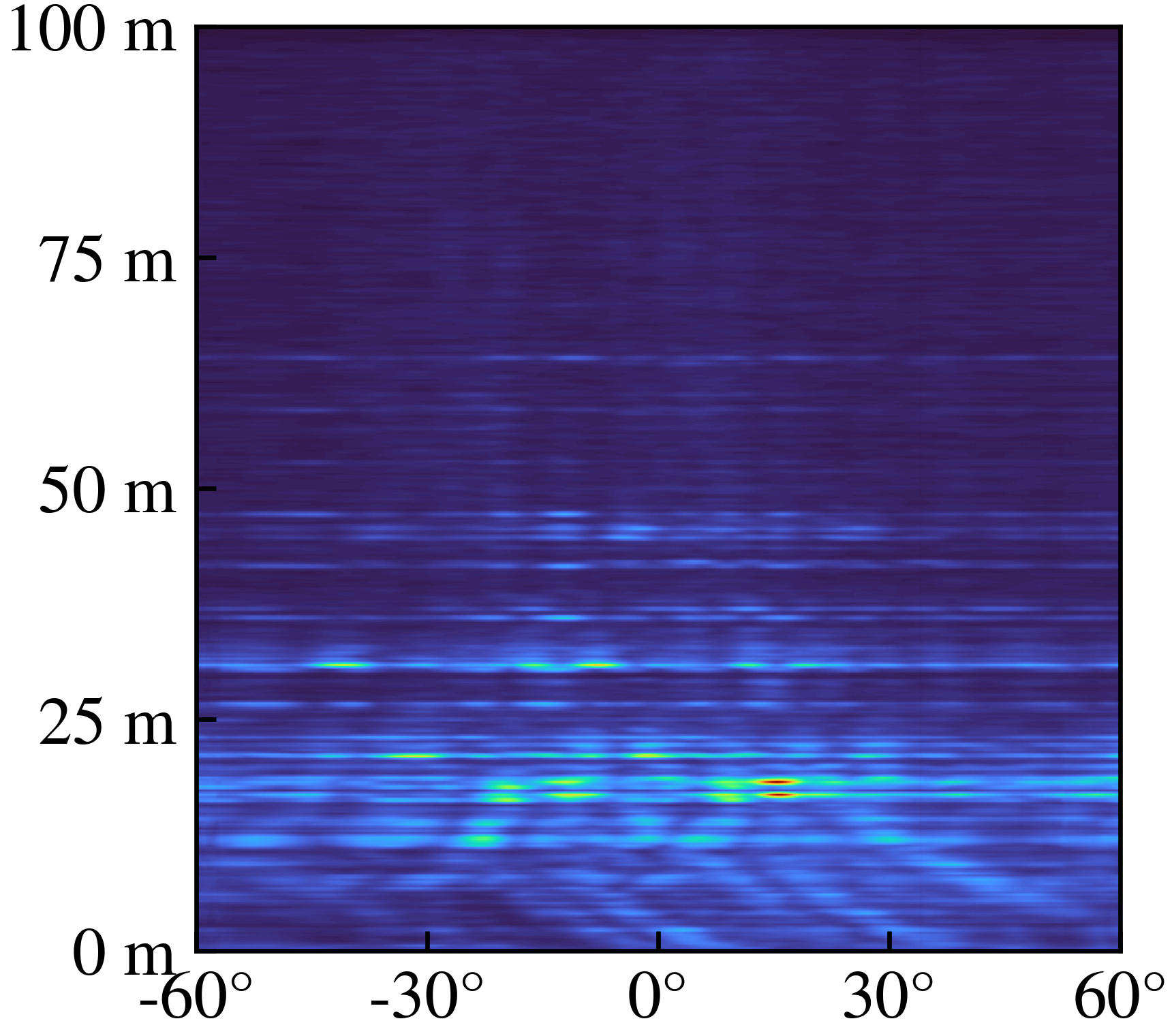}
    
    \caption{Effect of receiver reduction on MIMO radar range–azimuth (RA) maps. From left to right: 16~Rx, 7~Rx, and 5~Rx. The 7-Rx and 5-Rx layouts are obtained by uniformly subsampling the full 16-Rx array. All panels depict normalized received signal power under identical scaling.}
    \label{fig:intro_triptych}
\end{figure}

\section{Related Work}
\textbf{Deep Learning on Radar:}
Early radar perception methods use CFAR-based radar point clouds as the main input. 
These works adapt LiDAR-style 3D detectors to sparse radar measurements \cite{lang2019pointpillarsfastencodersobject,tauDelft2022}. 
RaTrack extends this paradigm to moving-object detection and tracking with point-wise scene flow and clustering \cite{pan2024ratrackmovingobjectdetection}. 
SemRaFiner addresses panoptic segmentation in sparse and noisy radar point clouds \cite{Zeller_2024}. 
All of these approaches rely on CFAR and clustering, which discard most amplitude, phase, and clutter information. 
Recent radar-only methods instead operate on raw radar tensors before CFAR. 
FFT-RadNet and T-FFTRadNet use complex range--Doppler spectra or ADC data for joint detection and freespace segmentation \cite{rebut2022,giroux2023tfftradnetobjectdetectionswin}. 
ADCNet learns directly from ADC data with embedded signal-processing blocks and distillation from offline RAD reconstruction \cite{yang2023adcnetlearningrawradar}. 
These methods show that raw radar yields better radar-only perception. 
However, they assume a fixed MIMO configuration and do not consider sensor fusion.
\par
\textbf{MIMO Radar Optimization and Antenna Design:} A separate line of work optimizes the MIMO array under sampling and hardware constraints. 
Classical compressed-sensing designs use sparse or random antenna layouts and reconstruct RA or RAD maps from sub-Nyquist measurements \cite{CS2006,Yu_2010,Rossi_2014}. 
These methods decouple sampling-pattern design from reconstruction. 
Recent approaches jointly learn the sampling pattern and the reconstruction network. 
Weiss et al.\ model MIMO sub-sampling, beamforming, and RA reconstruction in a single differentiable pipeline \cite{weiss2021mimoOptim}. 
Their method uses a relaxed multivariate Bernoulli parameterization and the Gumbel-Softmax distribution to select a subset of ADC channels within a fixed Rx budget \cite{jang2017gumbel,ronneberger2015unet}. 
Oral et al.\ extend this idea to sparse 2D MIMO arrays and near-field 3D imaging \cite{oral2025jointSparseMIMO}. 
These works show that optimal layouts depend on the reconstruction algorithm. 
They remain radar-only, assume a fixed Rx budget, and ignore fusion paradigms with other modalities.  
\par
\textbf{Multi-Modal Sensor Fusion with Radar:}
Beyond radar only pipelines, modern autonomous driving systems increasingly rely on BEV-based multi-modal fusion. 
BEVFusion unifies multi-view camera and LiDAR features in a shared BEV space with a view-transformation operator \cite{liu2024bevfusion}. 
MetaBEV extends this idea with meta-BEV queries and mixture-of-experts to handle missing or corrupted modalities \cite{ge2023metabev}. 
Several works incorporate radar by lifting point detections to 3D. 
CenterFusion \cite{Nabati_2021} and GRIF-Net \cite{kim2020grifnet} perform mid-level radar–camera fusion in the image plane, using radar points to refine depth or leveraging gated region-of-interest fusion mechanisms. 
In the LiDAR domain, RaLiBEV fuses LiDAR pillars and radar RA heatmaps \cite{yang2024ralibevradarlidarbev}, while InterFusion \cite{wang2022interfusion} proposes interaction-based fusion to handle sparse 4D radar data effectively. DeepFusion \cite{drews2022deepfusion} further demonstrates that modular fusion architectures are essential for robustness against sensor degradation. 
Recent LiDAR-radar studies show that naive fusion can underperform strong LiDAR-only baselines \cite{chae2024}. 
BEVGuide generalizes BEV-space fusion to arbitrary sensor sets using BEV-guided queries and cross-modal attention \cite{bevguide2023,li2023delvingdevilsbirdseyeviewperception}.  EchoFusion \cite{liu2023echoes} likewise employs a transformer-style cross-attention and operates directly on raw radar data. It exploits the fact that 3D points along a polar ray project to a single image column and a single row in the radar tensor. 
Polar-aligned BEV queries are initialized on a discrete grid of azimuth and range. 
Multi-level image and radar features are extracted by modality-specific backbones. 
Column-wise attention over image features and row-wise cross-attention over radar features let each BEV query aggregate information from its corresponding image column and radar row. 
A polar BEV decoder then refines these queries to predict 3D bounding boxes.
This perspective is consistent with prior work showing that camera fusion can compensate radar's limited angular resolution, particularly at longer ranges \cite{prabhakara2022mmwave}.

Existing BEV-based fusion methods either fuse cameras with LiDAR and radar point clouds, or incorporate raw radar as an additional modality while keeping the radar front-end fixed. 
In all cases, the radar configuration is predetermined and independent of the downstream fusion stack.
\par
In summary, existing radar–camera–LiDAR fusion methods treat the radar front-end as fixed, and existing MIMO design frameworks optimize only for radar-only reconstruction under a prescribed Rx budget. None of these approaches considers how the optimal radar configuration changes when radar is fused with other modalities. DeeperRadar fills this gap by learning both the antenna layout and the effective Rx budget directly from the multi-modal 3D detection task, with supervision from cameras and LiDAR. This yields sensor–stack-dependent radar designs.

\section{Method}
\label{sec:method}

We present an overview of the proposed DeeperRadar pipeline and its main components in Sec.~\ref{sec:method_overview}. 
Sec.~\ref{sec:radar_branch} details the radar branch, including the differentiable MIMO design module and antenna selection. 
Next in Sec.~\ref{sec:fusion}, we describe the multi-modal BEV fusion network and detection head. 
\subsection{Overview of the DeeperRadar Pipeline}
\label{sec:method_overview}

\begin{figure*}[t]
    \centering
    \includegraphics[width=\textwidth,height=0.27\textheight]{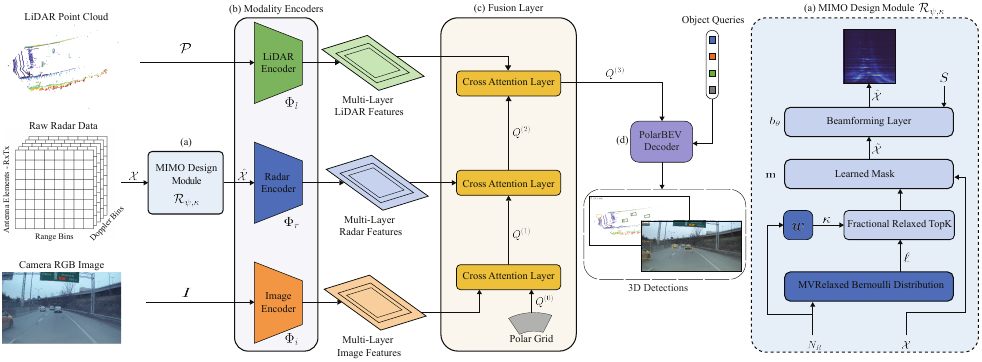}
    \caption{Overview of the proposed DeeperRadar pipeline. 
LiDAR point clouds, raw radar data, and camera images serve as input. 
(a) A MIMO design module learns a subsampling mask over radar receivers and produces subsampled range–azimuth maps. 
(b) Modality-specific encoders lift all inputs to multi-level BEV features. 
(c) Polar-aligned cross-attention fuses the BEV features into object queries, which (d) a PolarBEV decoder converts into 3D detections. 
The right panel shows an exploded view of the MIMO design module.}
    \label{fig:wide_chart}
\end{figure*}

Figure~\ref{fig:wide_chart} provides an overview of the proposed DeeperRadar pipeline. The architecture is organized into four main components: (a) a budget-aware radar branch with a MIMO design module that jointly learns the active receiver subset and budget, (b) modality-specific feature extraction backbones for LiDAR, radar, and cameras, (c) a polar-aligned cross-attention fusion layer, extended with an additional cross-attention path to incorporate the third modality, and (d) a multi-scale Polar BEV decoder as the detection head. The model is trained end-to-end on raw radar ADC tensors, camera RGB images, and LiDAR point clouds. At deployment, the MIMO design module is stripped away and replaced by the learned sparse Rx mask, which is applied directly to the radar ADC tensor to zero out the data corresponding to the removed antennas. In this way, the effective radar configuration is updated while the depth and computational cost of the downstream architecture remain unchanged compared to the baseline without the design module.

\subsection{Radar Branch and MIMO Design}
\label{sec:radar_branch}
We design the radar front-end by learning a sparse activation mask over the $N_R$ physical Rx-- channels (shared across all Tx). This mask selects a reduced virtual array. Concretely, a MIMO design module $\mathcal{R}_{\psi,\kappa}$ produces a sparse mask $\mathbf{m}$ that is applied to the raw ADC tensor
$\mathcal{X} \in \mathbb{C}^{N_{\text{chirp}}\times N_{\text{range}}\times D}$,
where $N_{\text{chirp}}$, $N_{\text{range}}$, and $D$ denote the numbers of chirps, range bins, and virtual channels, respectively.

Applying this mask yields a sub-sampled tensor 
$\tilde{\mathcal{X}} = \mathcal{X} \odot \mathbf{m} \in \mathbb{C}^{N_{\text{chirp}}\times N_{\text{range}}\times d}$, 
which is then beamformed by $b_\theta(\cdot)$ using a predefined steering matrix $S$ that accounts for per-antenna phase delays. This produces a subsampled RA map
$\hat{\mathcal{X}} = b_\theta(\tilde{\mathcal{X}}; S) \in \mathbb{R}^{N_\theta \times N_{\text{range}}}$,
where $N_\theta$ is the number of the angular bins. 
Here $D = N_{\text{R}}N_{\text{T}}$ and $d = n_{\text{R}}N_{\text{T}}$, with $N_{\text{R}}$ and $N_{\text{T}}$ being the numbers of Rx and Tx antennas in the full virtual array, and $n_{\text{R}} \leq N_{\text{R}}$ the learned effective number of active Rx.
\par
Following \cite{weiss2021mimoOptim,multigumbalsoftmax2020}, we treat $\mathbf{m}$ as a sample from a multivariate relaxed Bernoulli distribution on \((0,1)^{N_{\text{R}}}\). Concretely, we introduce a Gaussian copula over the antennas by sampling multivariate Gaussian vector
\begin{equation}
    \ g \sim \mathcal{N}(\mathbf{0}, \boldsymbol{\Sigma}), \qquad 
    \boldsymbol{\Sigma} = \mathbf{L}\mathbf{L}^{\top} \succeq \mathbf{0},
\end{equation}
and mapping it component-wise to correlated uniform random variables
\begin{equation}
    u_r \;=\; \Phi(g_r), 
    \qquad \sigma_r^2 = \Sigma_{rr}, 
    \qquad r = 1,\dots,N_{\text{R}},
\end{equation}
where $\Phi$ denotes the Gaussian CDF with variance \(\sigma_r^2\). These uniform random variables are then transformed into logistic scores
\begin{equation}
    \ell_r \;=\; \log \alpha_r + \log u_r - \log\!\bigl(1-u_r\bigr), 
    \qquad r = 1,\dots,N_{\text{R}},
\end{equation}
with learnable location parameters $\boldsymbol{\alpha}$ and a learnable covariance factor $\mathbf{L}$.
The vector \(\boldsymbol{\ell}\) serves as a correlated set of channel scores that feeds a continuous subset-selection operator to produce the soft mask $\mathbf{m}$.
\par
In contrast to \cite{weiss2021mimoOptim}, where the number $n_R$ of active Rx antennas is fixed a priori, we propose to tune this parameter in the course of learning. For this purpose, we additionally introduce a budget parameter \(\kappa\), implemented as a learnable weight $w$ that scales the total number of receive channels,
\begin{equation}
    \kappa \;=\; w\,N_{\text{R}},
\end{equation}
so that \(\kappa\) represents a continuous Rx budget. To obtain a soft selection mask consistent with this budget, we extend the relaxed Top-\(K\) operator \(\mathcal{T}_{K}\) of \cite{xie2021reparameterizablesubsetsamplingcontinuous} to the fractional setting and apply it to the scores \(\boldsymbol{\ell}\):
\begin{equation}
    \mathbf{m} \;=\; \mathcal{T}_{\kappa}(\boldsymbol{\ell}) \in [0,1]^{N_{\text{R}}}, 
    \qquad \mathbf{1}^{\top}\mathbf{m} \;=\; \kappa.
\end{equation}
We decompose
\begin{equation}
    \kappa \;=\; \lfloor \kappa \rfloor + \delta, 
    \qquad \delta \in [0,1),
\end{equation}
and perform \(\lfloor \kappa \rfloor\) standard relaxed Top-\(1\) iterations to allocate unit mass to the highest-scoring channels, followed by one additional iteration weighted by \(\delta\) to distribute the remaining fractional mass. This construction yields a fractional Top-\(K\) mask \(\mathbf{m}\) whose total selection mass exactly matches the continuous budget \(\kappa \approx n_{\text{R}}\). As a result, both the subset of active antennas (through the multivariate relaxed Bernoulli scores \(\boldsymbol{\ell}\)) and the effective number of selected channels (through \(\kappa\)) are learned jointly in a fully differentiable manner.
\par
We update the final loss function as follows:
\begin{equation}
\label{eq:loss}
    L = L_\text{task} + \lambda \kappa
\end{equation}
where $\lambda$ biases the solution toward sparser $n_R$ designs.

\subsection{Multi-Modal BEV Fusion Network}
\label{sec:fusion}
Given the subsampled radar RA map $\hat{\mathcal{X}}$ from Sec.~\ref{sec:radar_branch}, synchronized camera images $I$, and LiDAR point clouds $\mathcal{P}$, we construct a shared polar BEV representation through modality-specific encoders followed by sequential polar aligned cross-attention fusion.

Each modality is processed by a deep backbone 
followed by an FPN~\cite{lin2017featurepyramidnetworksobject}, yielding multi-scale feature maps:
\begin{align}
    \Phi_r(\hat{\mathcal{X}}) &= \{F_r^{(k)}\}_{k=1}^K = \mathrm{FPN}_r\bigl(\phi_r(\hat{\mathcal{X}})\bigr), \\
    \Phi_i(I) &= \{F_i^{(k)}\}_{k=1}^K = \mathrm{FPN}_i\bigl(\phi_i(I)\bigr), \\
    \Phi_l(\mathcal{P}) &= \{F_l^{(k)}\}_{k=1}^K = \mathrm{FPN}_l\bigl(\phi_l(\mathcal{P})\bigr),
\end{align}
where $\phi_r$, $\phi_i$, and $\phi_l$ denote the radar, image, and LiDAR encoder backbones, respectively.

Following~\cite{liu2023echoes,jiang2023polarformermulticamera3dobject}, we initialize learnable polar BEV queries $Q^{(0)} \in \mathbb{R}^{N_q \times C}$ arranged on a discrete grid of azimuth angles and radial distances. We apply sequential polar-aligned cross-attention blocks $\mathcal{A}_\cdot$ in image-radar-LiDAR order:
\begin{align}
    Q^{(1)} &= \mathcal{A}_i\bigl(Q^{(0)}, \{\Phi_i(I)\}_{x_I}\bigr), \label{eq:img_fusion} \\
    Q^{(2)} &= \mathcal{A}_r\bigl(Q^{(1)}, \{\Phi_r(\hat{\mathcal{X}})\}_{r_R}\bigr), \label{eq:radar_fusion} \\
    Q^{(3)} &= \mathcal{A}_l\bigl(Q^{(2)}, \Phi_l(\mathcal{P})\bigr), \label{eq:lidar_fusion}
\end{align}
where $\mathcal{A}_i$ attends to the \emph{image column features} $\{F_i^{(k)}(x_I)\}_{k=1}^K$. $\mathcal{A}_r$ attends to the corresponding \emph{range row features} $\{F_r^{(k)}(r_R)\}_{k=1}^K$ in the radar RA map and $\mathcal{A}_l$ performs aligned attention on LiDAR features across all pyramid levels $k=1,\dots,K$.

The final multi-modal BEV features $Q^{(3)}$ feed a polar BEV decoder~\cite{jiang2023polarformermulticamera3dobject} for 3D bounding box prediction. For single and two-modality ablation studies, the unused modality's FPN and corresponding cross-attention layer in Eqs.~(\ref{eq:radar_fusion}--\ref{eq:lidar_fusion}) are pruned.

\begin{table*}[htbp]
\caption{3D detection performance (AP@0.5) as a function of the fixed radar receiver budget $n_R$,
comparing heuristic subsampling with our learned antenna selection across three configurations:
\textit{Radar}, \textit{Radar+Camera}, and \textit{Radar+LiDAR}.
All values are reported as mean $\pm$ standard deviation over 9 training seeds.
For each setting, we report overall performance and a range split into 0--50\,m and 50--100\,m.
\textbf{Bold} indicates the better method (Learned vs. Heuristic) at a fixed $n_R$.
\highlight{Highlighted rows correspond to the learned optimal budget; for \textit{Radar}-only, the learned optimum is $n_R^\star=8$, which is not explicitly evaluated, so we highlight the nearest evaluated budget ($n_R=9$).}}
    \centering
  \vspace{0.35em}
    \setlength{\tabcolsep}{4pt}
    \renewcommand{\arraystretch}{1.1}
    \resizebox{\textwidth}{!}{
    \begin{tabular}{@{}C{0.4cm}c ccc|ccc|ccc@{}}
         \hline
        & & \multicolumn{3}{c}{\textbf{Radar}}
          & \multicolumn{3}{c}{\textbf{Radar + Camera}} 
          & \multicolumn{3}{c}{\textbf{Radar + LiDAR}} \\
        \cmidrule(lr){3-5} \cmidrule(lr){6-8} \cmidrule(l){9-11}
        $n_R$ & Method
        & Overall & 0--50\,m & 50--100\,m
        & Overall & 0--50\,m & 50--100\,m
        & Overall & 0--50\,m & 50--100\,m \\
        \midrule

        \multirow{2}{*}{3}
        & Heuristic
        & 24.98 $\pm$ 1.68 & 39.89 $\pm$ 2.74 & 9.41 $\pm$ 2.19
        & 44.91 $\pm$ 1.43 & 62.17 $\pm$ 2.53 & 27.09 $\pm$ 2.01
        & \highlight{50.01 $\pm$ 1.86} & \highlight{\textbf{67.75 $\pm$ 1.70}} & \highlight{32.25 $\pm$ 1.88} \\
        & Learned
        & \textbf{25.48 $\pm$ 1.43} & \textbf{40.44 $\pm$ 2.72} & \textbf{10.25 $\pm$ 1.32}
        & \textbf{46.01 $\pm$ 0.75} & \textbf{63.68 $\pm$ 0.97} & \textbf{27.30 $\pm$ 0.30}
        & \highlight{\textbf{50.28 $\pm$ 1.17}} & \highlight{67.50 $\pm$ 1.39} & \highlight{\textbf{34.43 $\pm$ 1.42}} \\
        \midrule

        \multirow{2}{*}{5}
        & Heuristic
        & 29.64 $\pm$ 1.40 & 44.39 $\pm$ 1.33 & 14.57 $\pm$ 2.07
        & 44.12 $\pm$ 1.90 & 61.12 $\pm$ 3.79 & 28.05 $\pm$ 1.68
        & 49.28 $\pm$ 2.84 & 67.24 $\pm$ 4.03 & 31.99 $\pm$ 1.35 \\
        & Learned
        & \textbf{30.33 $\pm$ 0.85} & \textbf{45.71 $\pm$ 1.70} & \textbf{14.77 $\pm$ 0.31}
        & \textbf{45.90 $\pm$ 2.06} & \textbf{62.24 $\pm$ 1.10} & \textbf{28.80 $\pm$ 3.32}
        & \textbf{50.78 $\pm$ 0.97} & \textbf{68.08 $\pm$ 0.86} & \textbf{33.52 $\pm$ 2.05} \\
        \midrule

        \multirow{2}{*}{7}
        & Heuristic
        & 29.83 $\pm$ 1.73 & 44.54 $\pm$ 2.11 & 15.07 $\pm$ 1.63
        & \highlight{45.00 $\pm$ 2.30} & \highlight{62.56 $\pm$ 4.40} & \highlight{28.38 $\pm$ 1.44}
        & 49.90 $\pm$ 1.86 & 68.04 $\pm$ 1.48 & 32.50 $\pm$ 3.56 \\
        & Learned
        & \textbf{31.00 $\pm$ 1.36} & \textbf{46.47 $\pm$ 1.92} & \textbf{15.49 $\pm$ 1.05}
        & \highlight{\textbf{47.63 $\pm$ 0.85}} & \highlight{\textbf{65.54 $\pm$ 2.48}} & \highlight{\textbf{31.00 $\pm$ 1.43}}
        & \textbf{50.31 $\pm$ 0.65} & \textbf{68.30 $\pm$ 1.70} & \textbf{33.42 $\pm$ 0.57} \\
        \midrule

        \multirow{2}{*}{9}
        & Heuristic
        & \highlight{32.05 $\pm$ 0.85} & \highlight{46.83 $\pm$ 1.27} & \highlight{\textbf{17.36 $\pm$ 1.17}}
        & 46.40 $\pm$ 1.24 & 63.83 $\pm$ 1.77 & 28.41 $\pm$ 1.33
        & 49.16 $\pm$ 2.55 & 68.24 $\pm$ 4.61 & 30.84 $\pm$ 2.43 \\
        & Learned
        & \highlight{\textbf{33.04 $\pm$ 0.59}} & \highlight{\textbf{48.40 $\pm$ 1.27}} & \highlight{17.35 $\pm$ 1.10}
        & \textbf{47.29 $\pm$ 1.04} & \textbf{63.91 $\pm$ 1.86} & \textbf{30.68 $\pm$ 0.78}
        & \textbf{50.67 $\pm$ 1.43} & \textbf{70.70 $\pm$ 1.25} & \textbf{31.91 $\pm$ 1.55} \\
        \midrule

        \multirow{2}{*}{12}
        & Heuristic
        & 32.26 $\pm$ 1.41 & 48.04 $\pm$ 1.58 & 16.52 $\pm$ 1.75
        & 47.13 $\pm$ 0.87 & 62.89 $\pm$ 1.49 & \textbf{30.92 $\pm$ 2.65}
        & 49.95 $\pm$ 1.50 & 68.04 $\pm$ 2.21 & 31.36 $\pm$ 2.79 \\
        & Learned
        & \textbf{33.67 $\pm$ 1.00} & \textbf{49.87 $\pm$ 1.80} & \textbf{17.34 $\pm$ 0.48}
        & \textbf{47.33 $\pm$ 2.33} & \textbf{64.89 $\pm$ 4.31} & 30.31 $\pm$ 0.98
        & \textbf{50.60 $\pm$ 1.80} & \textbf{69.45 $\pm$ 1.98} & \textbf{33.85 $\pm$ 1.72} \\
        \midrule

        16 & --
        & 33.67 $\pm$ 0.98 & 49.96 $\pm$ 0.74 & 17.65 $\pm$ 0.90
        & 45.22 $\pm$ 1.69 & 62.52 $\pm$ 2.59 & 28.91 $\pm$ 1.19
        & 50.25 $\pm$ 1.24 & 69.41 $\pm$ 1.60 & 32.74 $\pm$ 1.01 \\
        \bottomrule
    \end{tabular}
    }

\label{tab:detection_results_overall}
\end{table*}

\begin{figure*}[htbp]
    \centering

    \begin{minipage}{0.32\textwidth}
        \centering
        \textbf{\footnotesize\bfseries Radar}\\[0.3em]
        \includegraphics[width=\textwidth]{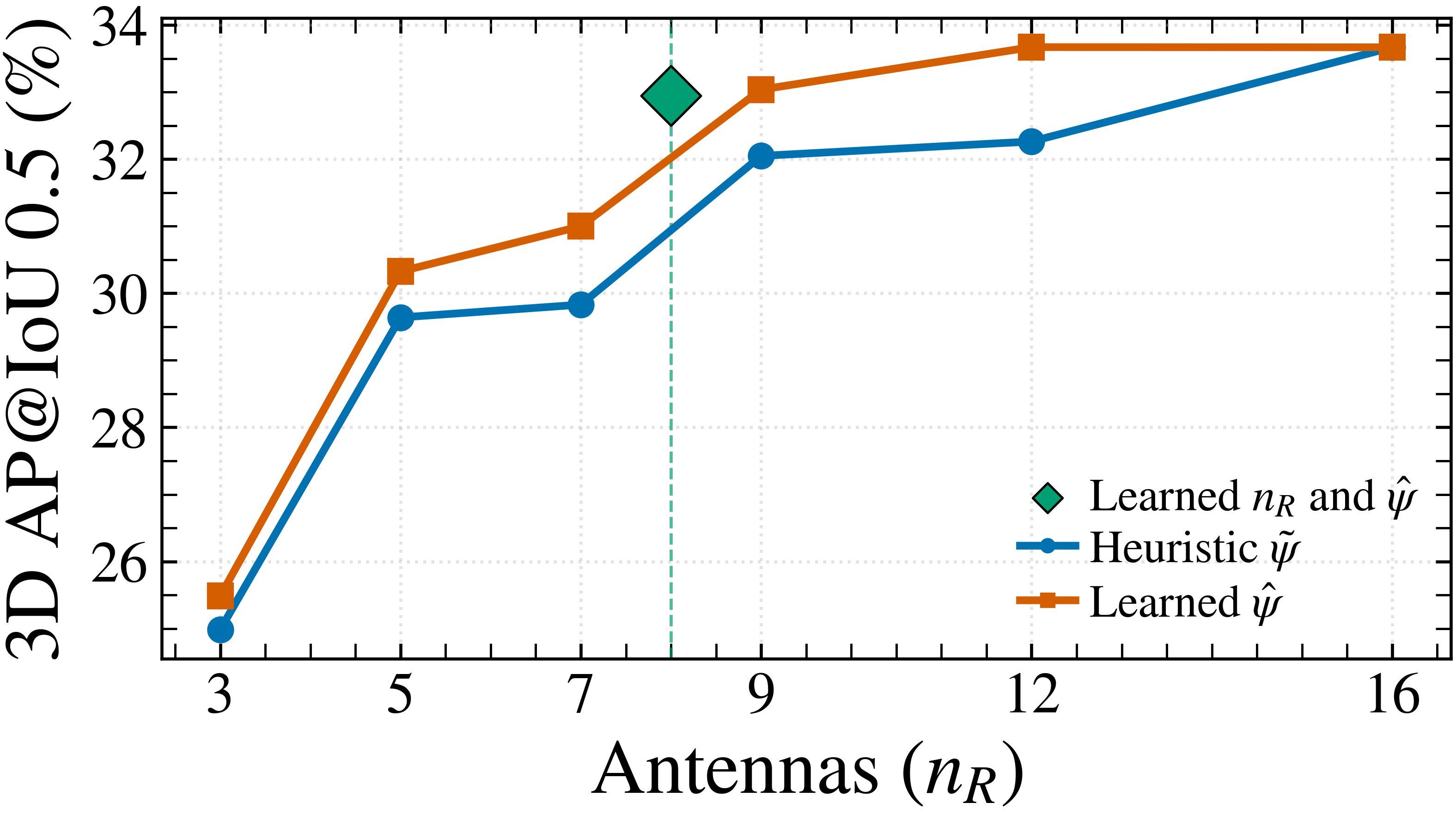}
    \end{minipage}
    \hfill
    \begin{minipage}{0.32\textwidth}
        \centering
        \textbf{\footnotesize\bfseries Radar + Camera}\\[0.3em]
        \includegraphics[width=\textwidth]{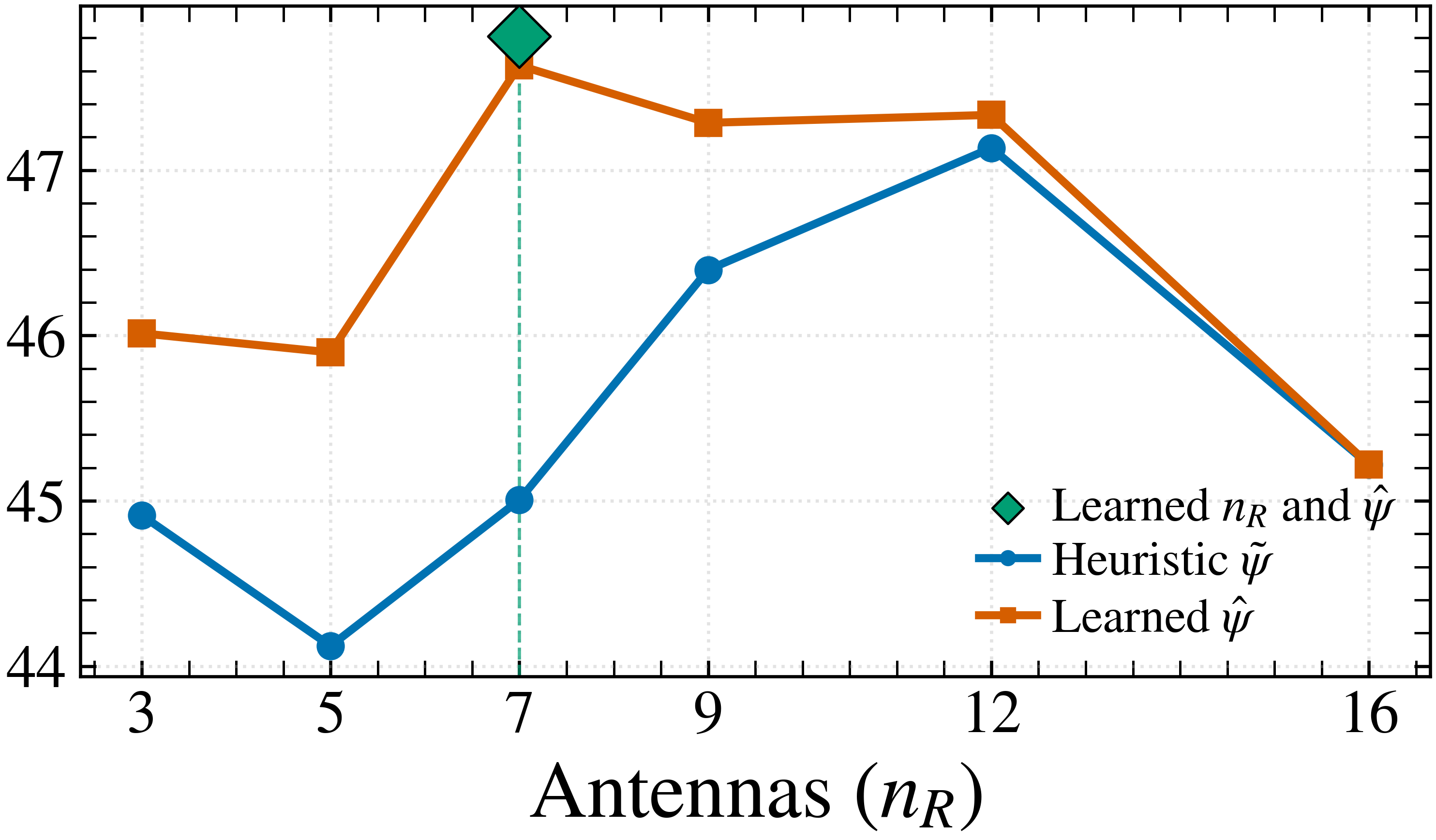}
    \end{minipage}
    \hfill
    \begin{minipage}{0.32\textwidth}
        \centering
        \textbf{\footnotesize\bfseries Radar + LiDAR}\\[0.3em]
        \includegraphics[width=\textwidth]{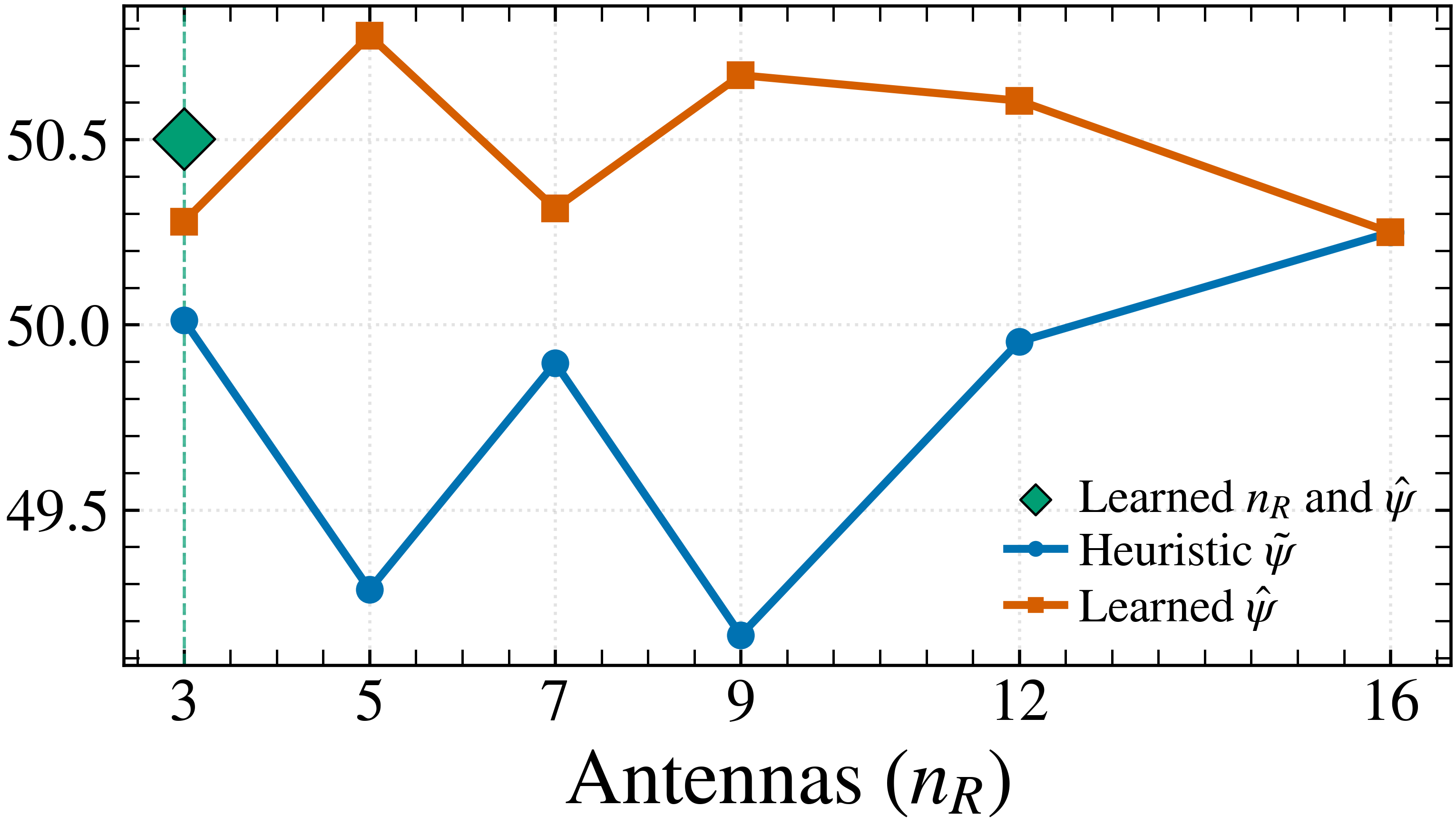}
    \end{minipage}

\caption{Overall 3D detection accuracy (AP@0.5) versus radar receiver budget $n_R$ for three configurations: \textit{Radar}, \textit{Radar+Camera}, and \textit{Radar+LiDAR}.
Each panel compares the mean performance of heuristic subsampling with the mean performance of learned antenna selection across random seeds.
The marker denotes the learned optimal budget $n_R^\star$ predicted by the budget-regularized objective.
These trends correspond to the overall metrics reported in Table~\ref{tab:detection_results_overall}; the radar-only panel serves as a complementary baseline evaluated without the fusion module.}
\label{fig:detection_results_overall_plots}
\end{figure*}

\section{Experiments}
Our experimental evaluation is organized around four key questions that structure the rest of this section: Subsection \ref{subsec:exp2} focuses on adapting and validating the learnable MIMO design in the automotive radar setting, while Subsection \ref{subsec:exp3} studies its impact on fusion-based 3D detection and receiver budget optimization. Concretely, we address the following questions:
\begin{enumerate}
    \renewcommand{\labelenumi}{Q\arabic{enumi})}
    \item Can the learnable MIMO design be adapted to the automotive radar domain?
    \item Can this design be integrated into a 3D detection pipeline with camera--LiDAR--radar fusion?
    \item Does our budget-aware model learn the optimal number and active subset of receive antennas?
    \item Can fusion compensate for reduced MIMO aperture and preserve or even improve 3D detection performance?
\end{enumerate}

\subsection{Experimental Setup}
\label{subsec:exp1}
We evaluate the proposed radar-centric, stack-conditioned MIMO design framework on two tasks: (a) RA reconstruction and (b) End-to-end 3D detection under different sensor stacks in the AV setting.

To adapt the MIMO design to the AV domain and validate our extended design module, we first run the original reconstruction task from \cite{weiss2021mimoOptim} on a simulated automotive radar dataset. The dataset was generated with a ray-tracing simulator \cite{radarsimx}, using driving scenes populated with 3D vehicle models from \cite{chang2015shapenetinformationrich3dmodel}. The simulator is configured with a MIMO radar similar to the RADIal sensor configuration of 16 Rx and 12 Tx antennas \cite{rebut2022}; since the exact RADIal radar configuration and parameters are not publicly available, we simulate a closely matching setup instead. From this simulated dataset, we use a 700/300 train–test splits and then apply the same reconstruction task to a similarly sized subset of real AV radar data from the RADIal dataset. Reconstruction quality is analyzed using the peak signal-to-noise ratio (PSNR) metric.
\par
We evaluate our proposed multi-modal end-to-end 3D detection MIMO design pipeline on the RADIal dataset using 3D average precision (AP) at an Intersection-over-Union (IoU) threshold of 0.5, reported overall and for two range intervals: 0--50\,m and 50--100\,m. We follow the annotation protocol provided in \cite{liu2023echoes}. The training split consists of 6471, and the test split consists of 1781 time-synchronized camera, radar, and LiDAR measurement frames.
\par
All experiments are run on a single NVIDIA RTX A6000 GPU. For the reconstruction task, we train for 200 epochs with a batch size of 10 using Adam \cite{adam2017} with learning rate $\eta_{\text{rec}} = 1\times 10^{-3}$. For the detection task, each model is trained for 24 epochs with a batch size of 1 using Adam with learning rate $\eta_{\text{det}} = 5\times 10^{-5}$ for the detection network and a separate learning rate $\eta_{\text{mimo}} = 3.5\times 10^{-5}$ for the MIMO design module. The regularization weight in the MIMO loss term~(\ref{eq:loss}) is fixed to $\lambda = 5\times 10^{-4}$ for all modality configurations. The learnable receiver budget is constrained to $2 \leq n_{\text{R}} \leq 16$, where $n_{\text{R}} = 16$ corresponds to the full Rx array.
\par
Across all experiments, we compare three configurations of our end-to-end pipeline:
(i) fixed $n_R \in \{3,5,7,9,12,16\}$ with heuristic initial Rx layouts $\tilde{\psi}$ from a fixed pool of 9 pre-generated layouts (3 uniform + 6 Sobol quasi-random \cite{dick_pillichshammer2010}), held fixed across runs; results are reported over 9 training seeds.
(ii) fixed $n_R$ with learned Rx subset $\hat{\psi}$; and
(iii) jointly learned Rx budget $n_R^\star$ and Rx subset $\hat{\psi}$.
\par
We report results for the \textit{Radar}-only, \textit{Radar+Camera}, and \textit{Radar+LiDAR} stacks. The \textit{Radar+Camera+LiDAR} setting exhibits higher run-to-run variance than the single- and two-modality stacks and is sensitive to the random seed.
We therefore report it as the best observed result across seeds in Table~\ref{tab:Fulldetresults3d}, rather than as part of the fixed-budget ablation study.

\subsection{Adapting Learnable MIMO Design to the AV Domain}
\label{subsec:exp2}
\begin{figure}[t]
    \centering
    \begin{minipage}{0.49\linewidth}
        \centering
        \textbf{\footnotesize\bfseries Simulation}\\[0.12em]
        \includegraphics[width=\linewidth]{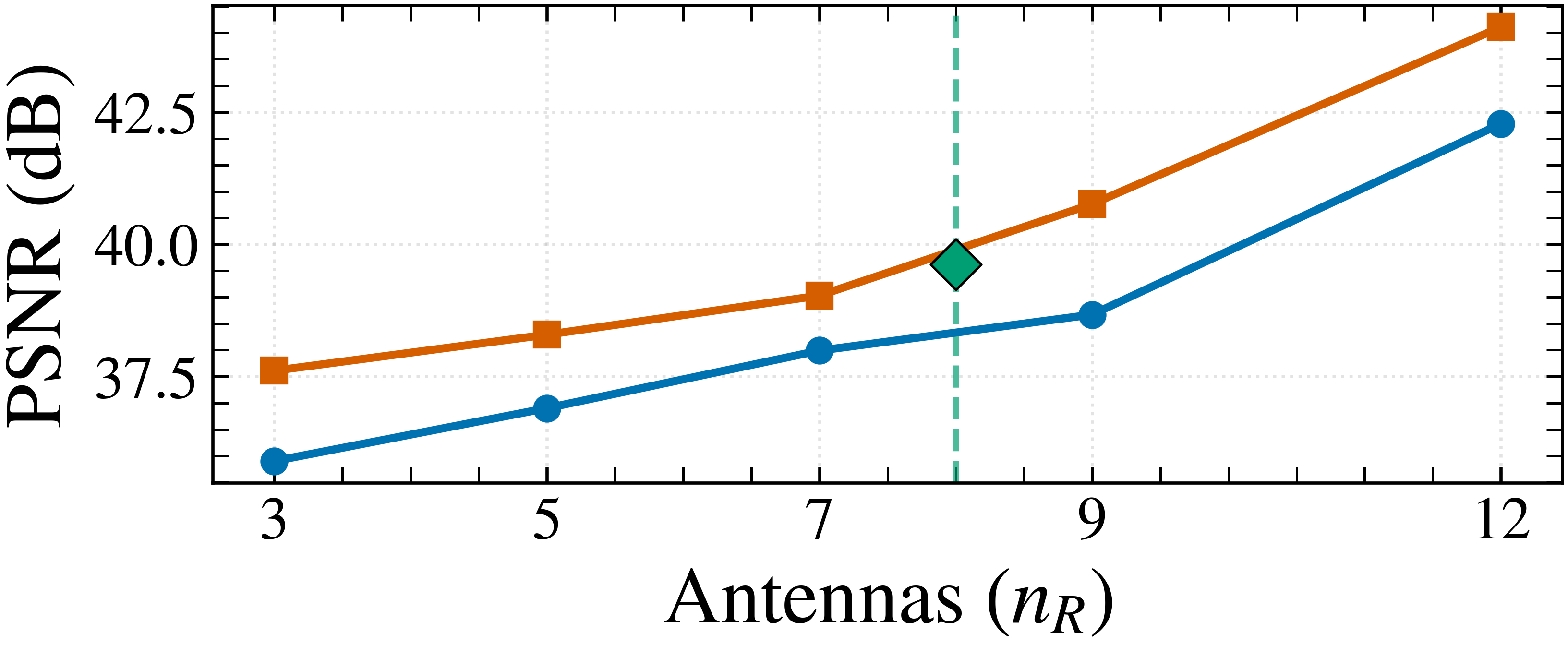}
    \end{minipage}\hfill
    \begin{minipage}{0.49\linewidth}
        \centering
        \textbf{\footnotesize\bfseries RADIal}\\[0.12em]
        \includegraphics[width=\linewidth]{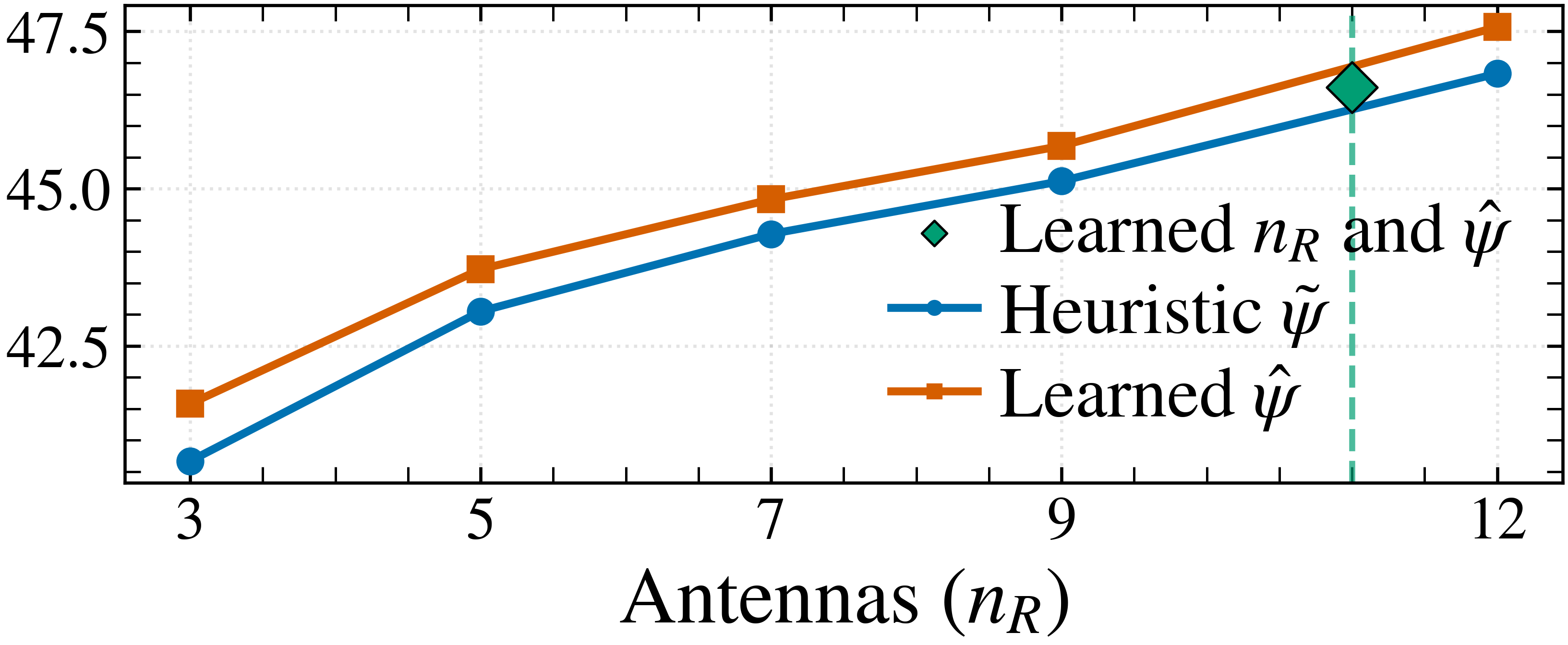}
    \end{minipage}
    \caption{Reconstruction PSNR versus number of antennas $n_R$ for simulation (left) and RADIal (right). Results are averaged over all random seeds. The marker denotes the learned optimal budget $n_R^\star$ predicted by the budget-regularized objective.}
    \label{fig:recon_psnr_sim_radial}
\end{figure}

To answer \textbf{Q1}, whether the learnable MIMO design transfers to automotive MIMO radar, we evaluate it on the RA reconstruction task in both simulation and real RADIal dataset. 
Figure~\ref{fig:recon_psnr_sim_radial} shows that, for every fixed budget $n_R$ and on both datasets, learned active Rx subsets $\hat{\psi}$ improve PSNR over heuristic Rx layouts. 
On average, learned layouts yield about a $4.2\%$ PSNR gain in simulation and $1.6\%$ in RADIal. 
For both datasets, PSNR increases monotonically with $n_R$. 
The learned budget extension, therefore, primarily acts as a design knob. 
By tuning the regularization weight $\lambda$ in~\eqref{eq:loss}, one can select a point with a favorable PSNR/$n_R$ trade-off. 
Specifically, the learned budget converges to $n_R^\star=8$ on the simulated dataset, achieving a PSNR of $39.62$ dB, and to $n_R^\star=11$ on the RADIal dataset, achieving $46.61$ dB.
Overall, optimizing which antennas are used yields better RA reconstruction than heuristic layouts on both simulated and real automotive MIMO radar data.

\subsection{Fusion-Aware MIMO Design for 3D Detection}
\label{subsec:exp3}
We now turn to questions \textbf{Q2--Q4} and study how the learnable MIMO design
$\mathcal{R}_{\psi,\kappa}$ affects downstream 3D detection under different sensor stacks. 
\par
\begin{figure}[t]
    \centering

    \begin{minipage}[c]{0.06\linewidth}
        \centering
        \rotatebox{90}{\footnotesize $\hat{\psi}$ 3D@IoU0.5 AP (\%)}
    \end{minipage}%
    \begin{minipage}[c]{0.90\linewidth}
        \centering
        \setlength{\tabcolsep}{2pt}
        \begin{tabular}{@{}ccc@{}}
            \begin{minipage}[t]{0.30\linewidth}
                \centering
                {\scriptsize \textbf{Radar}}\\[0.3em]
                \includegraphics[width=\linewidth]{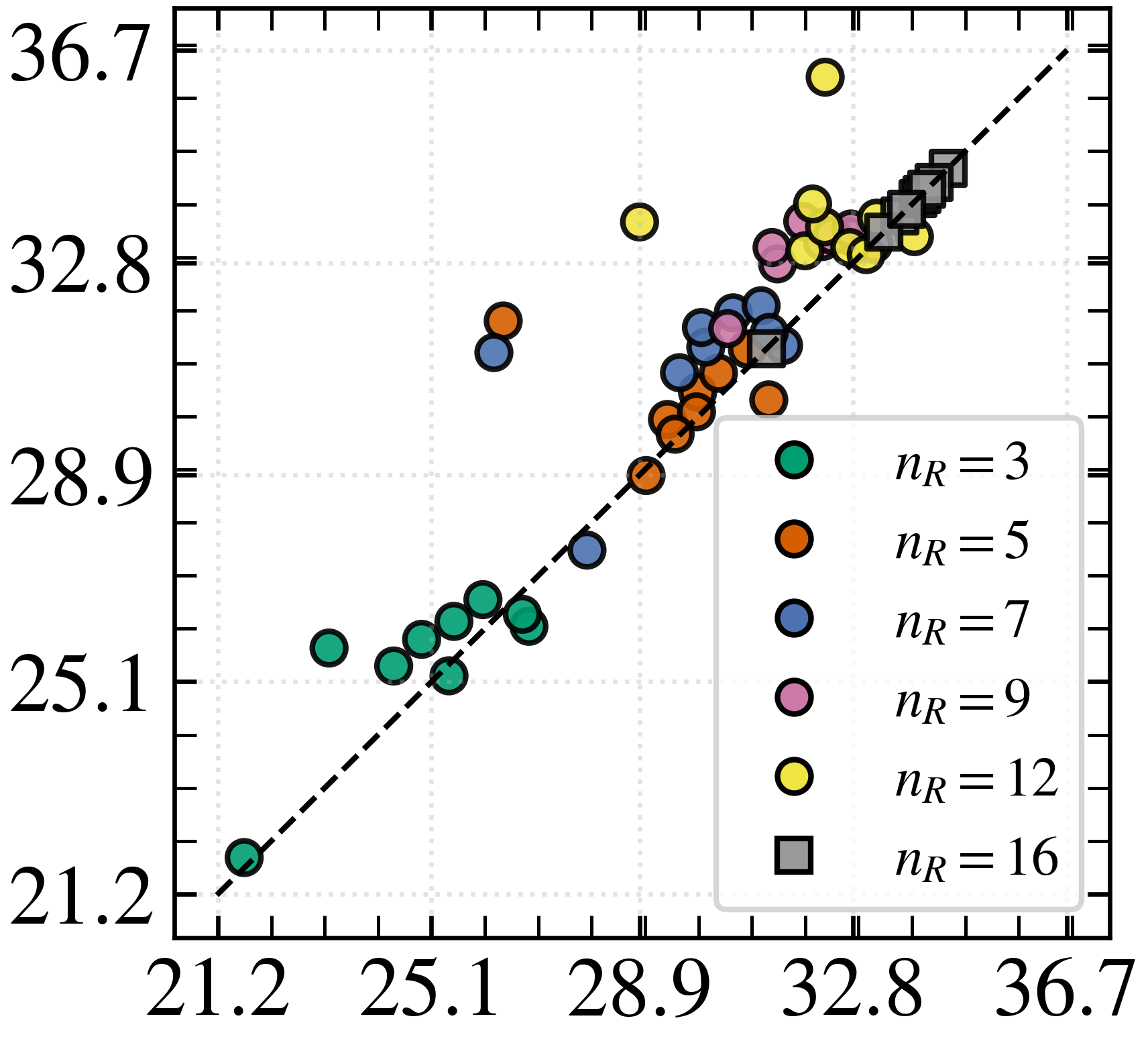}
            \end{minipage}
            &
            \begin{minipage}[t]{0.30\linewidth}
                \centering
                {\scriptsize \textbf{Radar + Camera}}\\[0.3em]
                \includegraphics[width=\linewidth]{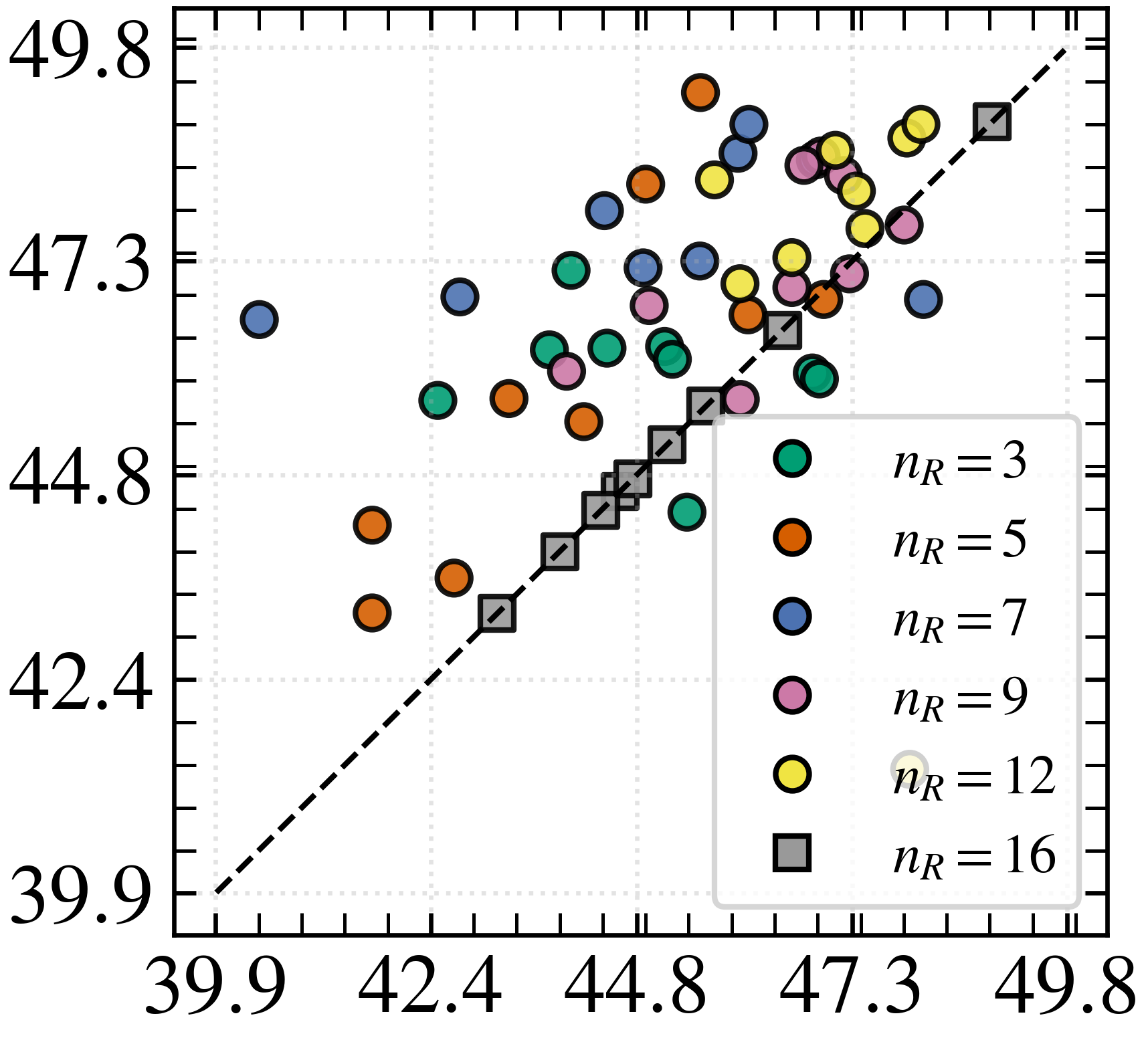}
            \end{minipage}
            &
            \begin{minipage}[t]{0.30\linewidth}
                \centering
                {\scriptsize \textbf{Radar + LiDAR}}\\[0.3em]
                \includegraphics[width=\linewidth]{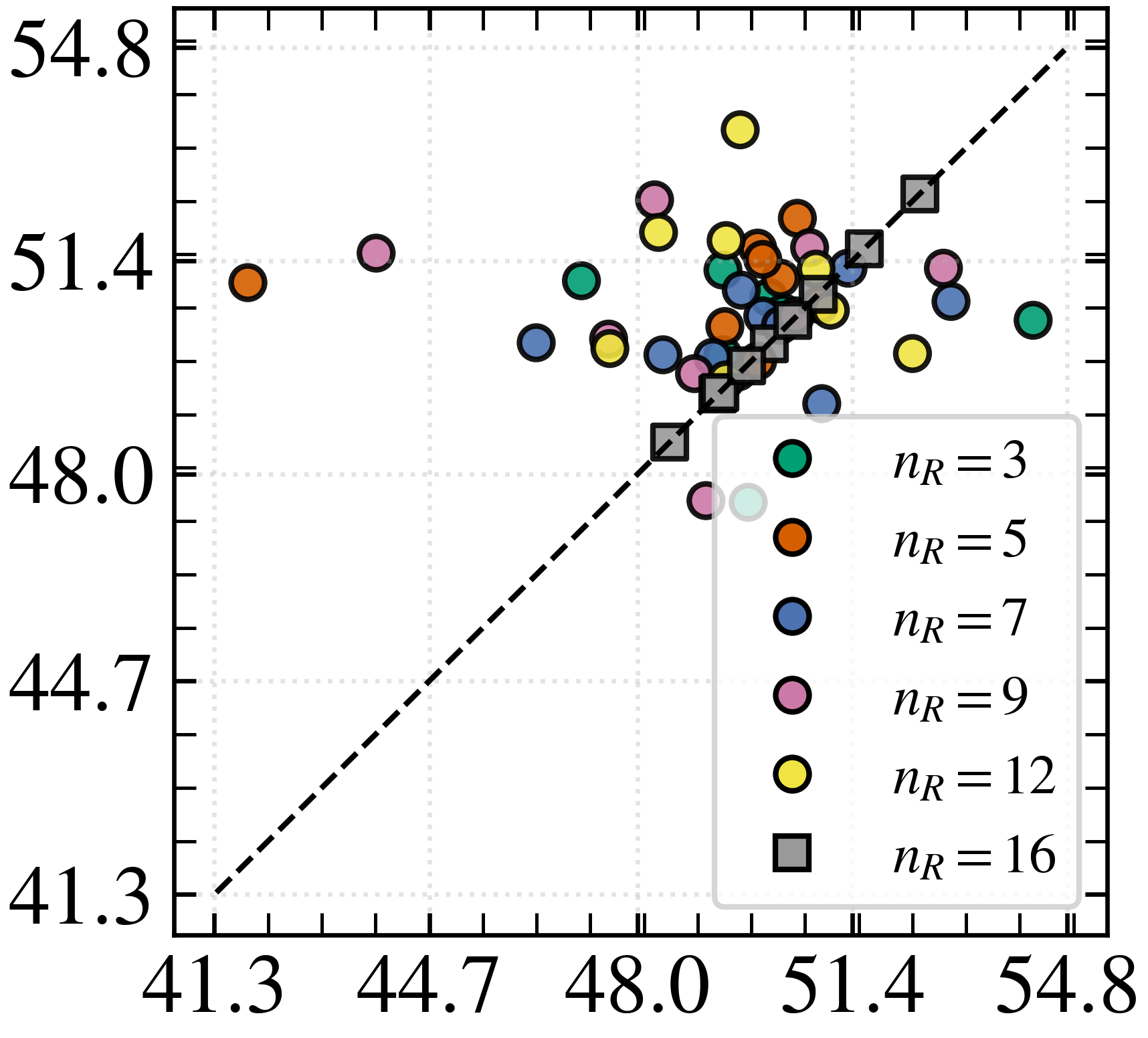}
            \end{minipage}
        \end{tabular}

        \vspace{0.35em}
        {\footnotesize $\tilde{\psi}$ 3D@IoU0.5 AP (\%)}
    \end{minipage}
    \caption{Comparison of learned layouts $\hat{\psi}$ versus fixed layouts $\tilde{\psi}$ across different sensor stacks. Each point represents one of 9 training seeds per configuration. The diagonal indicates parity between fixed and learned antenna selections in terms of 3D AP.}
    \label{fig:seed_compare}
\end{figure}

\textbf{Learning the Active Rx Subset $\hat{\psi}$ for Multi-Modal 3D Detection}:
For each fixed budget $n_R$, we compare heuristic Rx layouts $\tilde{\psi}$ to learned layouts $\hat{\psi}$ under the same detection architecture and sensor stack, and report the resulting 3D AP.
We run the radar-only, radar–camera, and radar–LiDAR, pipelines and summarize the per seed results in Fig.~\ref{fig:seed_compare} and overall results in Table~\ref{tab:detection_results_overall}. 
Across all sensor stacks and budgets, the learned layouts $\hat{\psi}$ achieve higher mean AP than the corresponding heuristic selections $\tilde{\psi}$, suggesting that the detector benefits from fusion-aware antenna placement.
In the \textit{Radar}-only setting, learned layouts improve overall AP over heuristic layouts by $+0.5$ to $+1.41$ points ($\approx 2\%$–$4\%$ relative), with the largest gain at $n_R=12$ (33.67 vs.\ 32.26 AP). Overall performance degrades as $n_R$ is reduced.
In the \textit{Radar+Camera} setting, learned layouts improve overall AP by $+0.20$ to $+2.63$ points for $n_R\!\in\!\{3,5,7\}$, corresponding to relative gains of $2\%-6\%$, with the strongest improvement at $n_R=7$ (47.63 vs.\ 45.00 AP). 
In the \textit{Radar+LiDAR} setting, learned layouts improve AP across all budgets (+0.27 to +1.51), with up to +8\% better long-range performance (e.g., 33.85 vs.\ 31.36 AP at $n_R=12$). Overall AP is relatively insensitive to $n_R$ in this stack (Fig.~4), suggesting diminishing performance from the MIMO radar when LiDAR is present.

\par
\textbf{Joint Learning of Active Rx Subset $\hat{\psi}$ and Budget $n_R^\star$}:
We next enable the budget parameter $\kappa$ and learn both the active Rx
subset and the effective Rx budget end-to-end using the
budget-regularized loss in~\eqref{eq:loss}.
Figure~\ref{fig:detection_results_overall_plots} visualizes overall 3D AP as a
function of $n_R$ for each sensor stack, with the green marker denoting the
learned budget $n_R^\star$.
\par
For the \textit{Radar}-only configuration, budget learning converges to $n_R^\star=8$.
This achieves $32.95$ overall AP (48.10 at 0--50m and 17.81 at 50--100m).
The full $n_R=16$ array reaches $33.67$ AP.
Thus, $n_R^\star=8$ is within $0.72$ AP while using 50\% fewer receivers.
Figure~\ref{fig:detection_results_overall_plots} shows a plateau beyond $n_R\approx 9$. Marginal gains are small for larger arrays.
\par
For the \textit{Radar+Camera} configuration, budget learning converges to
$n_R^\star = 7$, achieving an overall AP of $47.81$ (62.01 at 0--50\,m and
32.34 at 50--100\,m), compared to $45.22$ AP for the full $n_R=16$ array; this
corresponds to a gain of about $+2.6$ AP ($\approx 6\%$ relative) while using
56\% fewer receivers. 
\par
For the \textit{Radar+LiDAR} configuration, the model collapses the radar
front-end to a very sparse design with ${n}_R^\star = 3$, yet still reaches
$50.50$ overall AP (67.89 at 0--50\,m and 33.11 at 50--100\,m), slightly
improving over the full array ($50.25$ AP) with more than 80\% fewer Rx
channels. 
\par
Across all modality combinations, the overall detection curves exhibit non-negligible variance across Rx budgets and sensor stacks.
However, a deeper inspection of the per-experiment outcomes in Fig.~\ref{fig:seed_compare}, together with the aggregate trends in Fig.~\ref{fig:detection_results_overall_plots}, shows a consistent trend toward higher mean AP for learned Rx placements over heuristically chosen designs.
Moreover, in all fusion settings the best operating point occurs at a reduced budget $n_R^\star$, matching or slightly surpassing the full $n_R=16$ array.

\textbf{End-to-End 3D Detection with Fusion-Aware MIMO Design:}
We next compare the full DeeperRadar pipeline, with jointly learned $\hat{\psi}$ and $n_R^\star$, against non-MIMO-design baselines for each sensor stack. We establish baselines on the RADIal split using the original EchoFusion implementation~\cite{liu2023echoes} for single-modality camera, radar, LiDAR, and radar–camera fusion, and we retrain these models on our split for a fair comparison. We then train DeeperRadar with the proposed MIMO design module and evaluate it on the same data, summarizing the best overall 3D AP@0.5 results in Table~\ref{tab:Fulldetresults3d}. At inference time, all models use pipelines of identical depth; In DeeperRadar, the MIMO design module is removed and replaced by the learned sparse Rx mask $\mathbf{m}$, which is applied once to the raw ADC tensor.

In the radar–camera configuration, DeeperRadar achieves comparable or slightly higher overall 3D AP than the EchoFusion baseline while reducing the number of Rx antennas from 16 to 7, yielding an improvement in AP points at lower MIMO complexity. In the radar–LiDAR setting, fusion under performs a strong LiDAR-only baseline, in line with prior reports for simple cross-attention LiDAR–radar fusion~\cite{chae2024}.
In the radar–camera–LiDAR configuration, DeeperRadar achieves the best observed performance across our runs, reaching $71.95$ overall AP (81.61 at 0--50,m and 71.70 at 50--100,m) in Table~\ref{tab:Fulldetresults3d}.
\par
\begin{figure}[!t]
    \centering
    \makebox[\linewidth][c]{%
        \includegraphics[width=1\linewidth,height=1\textheight,keepaspectratio]{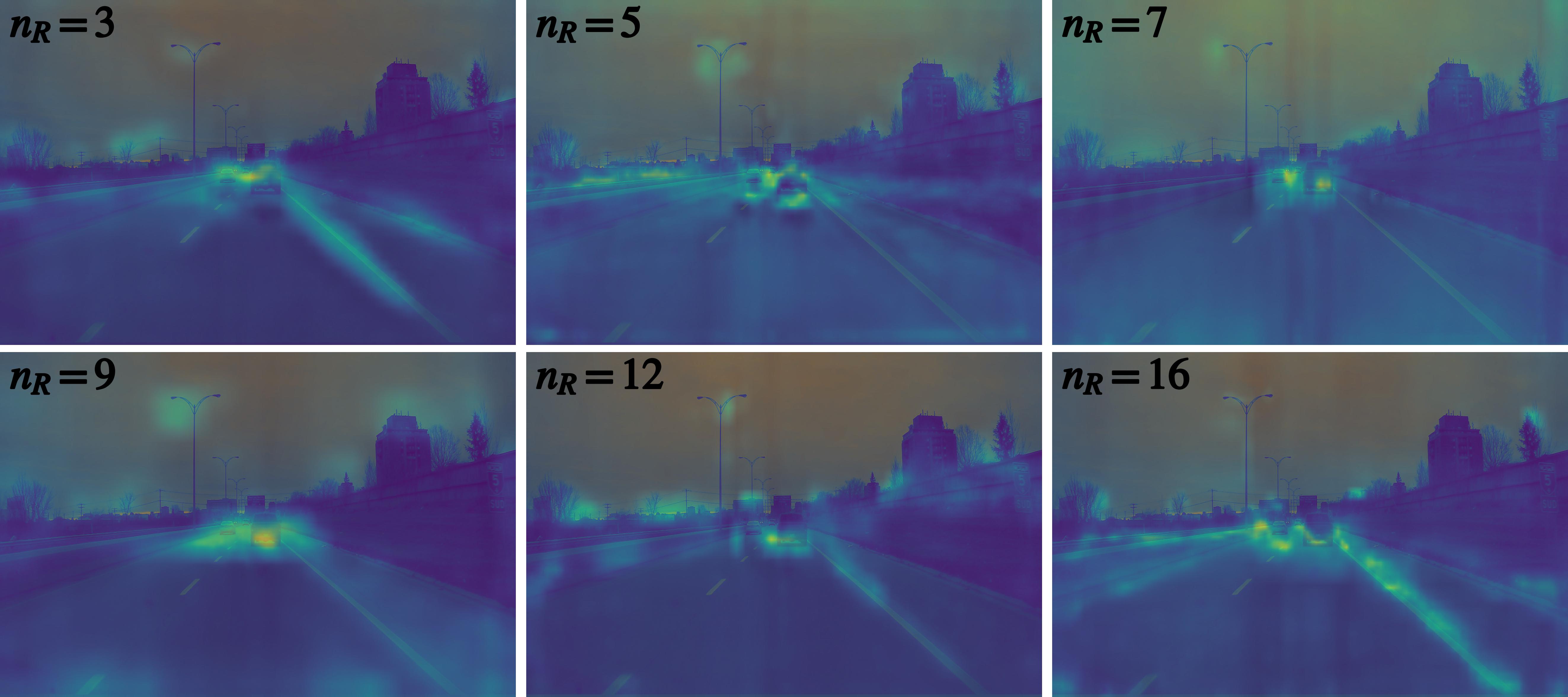}%
    }\\[0.35em]
    \includegraphics[width=1\linewidth,height=1\textheight,keepaspectratio]{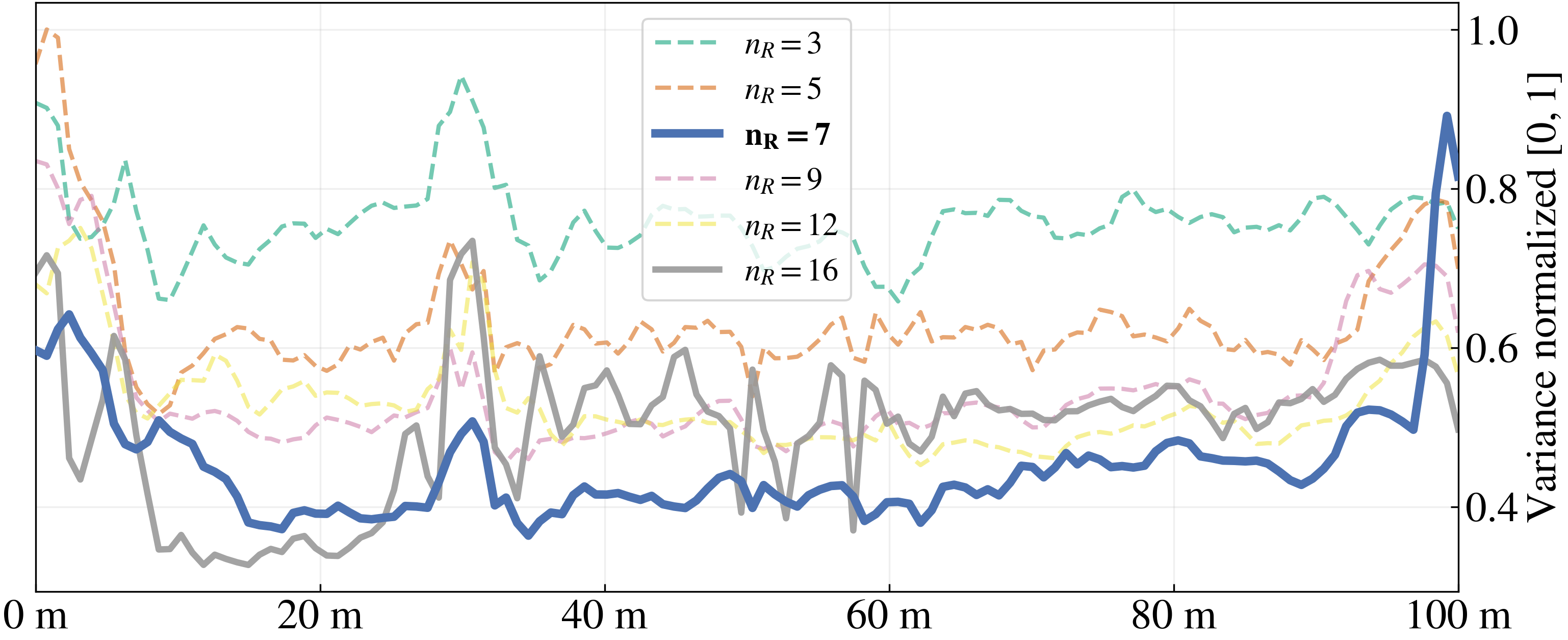}
    \caption{Attention analysis for varying radar budgets $n_R \in \{3,5,7,9,12,16\}$. 
    Top: $\mathcal{A}_i$ (\emph{which vertical slice of the image best explains each polar BEV cell}). 
    Bottom: per-range variance of $\mathcal{A}_r$ over azimuth (\emph{how strongly each range cell relies on radar azimuth bins}). 
    The best learned configuration, $n_R=7$, is highlighted in \textbf{bold}.}
    \label{fig:attention}
\end{figure}

We attempt to explain the learned fusion behavior, by visualizing the image-to-polar attention maps $\mathcal{A}_i$ and the radar-to-image attention maps $\mathcal{A}_r$ in Fig.~\ref{fig:attention}.
As we reduce $n_R$, the $\mathcal{A}_i$ maps shift from thin, edge-like responses for the full array ($n_R=16$) to compact, object-centered blobs near the learned optimum of $n_R$.
This evolution indicates that BEV queries increasingly rely on stable image semantics as the radar angular resolution coarsens.
\par
Furthermore, we plot the per-range variance of $\mathcal{A}_r$ across azimuth, which measures how strongly each range cell depends on the radar’s angular structure.
High variance corresponds to peaky, unstable azimuth responses, whereas low variance indicates smoother and more consistent aggregation of radar information.
The full array exhibits large, rapidly fluctuating variance, while very small arrays ($n_R=3$) lack clear angular definition.
The learned optimum at $n_R^\star=7$ yields the most stable variance profile across range.
Taken together, these patterns and the detection AP trends in Table \ref{tab:Fulldetresults3d}, suggest that moderate subsampling produces radar features that fuse more cleanly with the image, with angular reasoning increasingly delegated to the camera while radar continues to provide range cues that are largely preserved under antenna reduction.
\par

\begin{table}[!htbp]
\caption{3D detection performance (AP@0.5) on the RADIal dataset for different sensor stacks.
We compare EchoFusion baselines with DeeperRadar using the jointly learned active Rx subset and budget.
For each configuration, we report best overall AP and range-based splits.
The learned receiver budget is annotated with a superscript $\star$.
}
    \centering
    \begin{tabular}{lccccc}
        \multicolumn{6}{c}{} \\
        \hline
        Model & Modality & Overall & 0--50 m & 50--100 m & $n_R$ \\
        \hline
        EchoFusion    & C   &  6.36  & 16.93 & 0.46 & -  \\
        EchoFusion    & L   & 59.71  & 76.11  & 43.82 & -  \\
        EchoFusion    & R   & 33.25  & 48.40 & 18.10 & 16 \\
        EchoFusion    & R+C & 46.10  & 62.50 & 29.71& 16 \\      
        \hline
        DeeperRadar   & R   & 32.95  & 48.10  & 17.81 & $8^\star$ \\
        DeeperRadar   & R+C & 47.81    & 62.01   & 32.34& $7^\star$  \\
        DeeperRadar   & R+L &  50.50      & 67.89       & 33.11   & $3^\star$      \\
        DeeperRadar   & R+C+L & 71.95     & 81.61       & 71.70  & $7^\star$    \\
        \hline
    \end{tabular}

    \label{tab:Fulldetresults3d}
\end{table}

\section{Conclusion and Future Directions}
We introduced DeeperRadar, a radar-centric framework that jointly learns sparse MIMO receiver design and sensor-stack-conditioned BEV fusion for 3D detection from raw radar, camera, and LiDAR.
We evaluate the method on RADIal under multiple sensor stacks.
Learned receiver layouts improve 3D detection AP compared to heuristic layouts.
Learning the receiver budget alongside the layout reveals that the best operating point depends on the fusion stack.
In radar--camera fusion, the model prefers a reduced Rx budget and achieves higher AP than the full array while lowering MIMO complexity.
In the full radar--camera--LiDAR stack, DeeperRadar achieves the strongest observed results compared to available baselines.
Attention analysis supports a consistent mechanism: as radar angular resolution is reduced, the model shifts angular reasoning toward the camera while preserving radar’s range structure.
These findings show that array design is not a standalone radar problem, but a perception-driven choice shaped by the downstream task and the available complementary sensors.
\par
This work impacts autonomous perception by reframing radar as a co-designed component of the detection pipeline.
It provides an end-to-end way to trade active receiver count and integration complexity against task performance.
It suggests that “more antennas” is not a universal recipe once radar is fused with vision and LiDAR, motivating platform-specific radar configurations that match the sensor suite and compute envelope of the vehicle.
\par
Future work will extend the co-design framework beyond Rx selection to include Tx selection. Additional directions include Doppler-aware objectives, improved training stability, and robustness to domain shifts. Broader validation across additional datasets and hardware platforms will further support the deployment of DeeperRadar.



\bibliographystyle{IEEEtran}
\bibliography{references}
\addtolength{\textheight}{-12cm}

\end{document}